\pdfoutput=1

\documentclass[11pt]{article}

\usepackage[]{ACL2023}

\usepackage{times}
\usepackage{latexsym}

\usepackage{graphicx}
\usepackage[capitalize]{cleveref}
\usepackage{algorithm}
\usepackage{algpseudocode}
\usepackage{subfig}
\usepackage[T1]{fontenc}

\usepackage[utf8]{inputenc}

\usepackage{microtype}

\usepackage{inconsolata}

\usepackage{color,soul}

\title{Conformal Nucleus Sampling}

 \author{Shauli Ravfogel\textsuperscript{\normalfont1,2} \,  Yoav Goldberg\textsuperscript{\normalfont1,2}  \, Jacob Goldberger\textsuperscript{\normalfont 1}\\
\textsuperscript{1}Bar-Ilan University \, \textsuperscript{2}Allen Institute for Artificial Intelligence \\ 
  {\tt\{\href{mailto:shauli.ravfogel@gmail.com}{shauli.ravfogel}, \href{mailto:yoav.goldberg@gmail.com}{yoav.goldberg}\}@gmail.com} , \href{mailto:jacob.goldberger@biu.ac.il}{jacob.goldberger@biu.ac.il}
   }

\begin{document}
\maketitle
\begin{abstract}
Language models generate text based on successively sampling the next word. A decoding procedure based on nucleus (top-$p$) sampling chooses from the smallest possible set of words whose cumulative probability exceeds the probability $p$. 
In this work, we assess whether a top-$p$ set is indeed aligned with its probabilistic meaning in various linguistic contexts.
We employ conformal prediction, a calibration procedure that focuses on the construction of minimal prediction sets according to a desired confidence level, to calibrate the parameter $p$ as a function of the entropy of the next word distribution. We find that OPT models are overconfident, and that calibration shows a moderate inverse scaling with model size.
\newline
\newline
\vspace{1.5em} 
\hspace{.5em}\includegraphics[width=1.25em,height=1.25em]{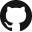}\hspace{.75em}\parbox{\dimexpr\linewidth-2\fboxsep-2\fboxrule}{\url{https://github.com/shauli-ravfogel/conformal-prediction}}
\end{abstract}

\section{Introduction}

Modern language generation methods are all based on  computing the conditional next-word distribution. However, there is still considerable debate about the best way to extract the next word from that distribution.
Most current text generation methods employ one of a handful of standard decoding strategies, which are  characterized as either deterministic or stochastic in nature.
A greedy search strategy selects the word with the highest probability at each timestep.
The greedy method and its beam search variations work remarkably well for machine translation but outside of this context, tend to return dull text or degenerate text \citep{holtzman2019curious,eldan2019}.
 \citet{holtzman2019curious} argued that high-quality human language does not follow a pattern of highest-probability next words, as humans expect the generated text to not be repetitive or boring. The same problem occurs with beam search. 
 
 Direct sampling from the next-word distribution computed by the model often generates incoherent gibberish text.
Temperature sampling \citep{ackley1985learning} is a word sampling approach based on rescaling logit scores before applying the softmax function to compute the word distribution.  
 Other methods limit the sampling space to a small \textbf{prediction set}
to avoid the ``unreliable tail''  \citep{holtzman2019curious}.
In top-$k$ sampling \citep{fan2018hierarchical}, we sample only from the top-$k$ most likely words. 
Instead of sampling only from the most likely $k$ words,  top-$p$ (nucleus) sampling chooses from the smallest possible set of words whose cumulative probability exceeds the probability $p$  \citep{holtzman2019curious}. Top-$p$ sampling  enables a dynamically sized window of words, unlike top-$k$ which fixes the size of $k$ for every step. 
Finally, locally typical sampling \citep{Meister2022} and truncation sampling  \citep{Hewitt2022} are recent variants of top-$p$
that aim to make it more suitable for language generation. 

The top-$p$ prediction set has a concrete probabilistic interpretation.
Here we examine  whether  the probability that the ``correct'' word belongs to the set of words produced by the top-$p$ algorithm is indeed $p$. 
More generally we expect that the next-word prediction would be calibrated, meaning that the output of the next-word softmax layer would accurately reflect the true word distribution. Parametric calibration methods, such as Temperature Scaling \cite{Guo2017}, which adjust the confidence of the most probable word, are not suitable for adjusting the size of the prediction set. 
Conformal Prediction (CP) ~\cite{vovk1999machine,vovk2005algorithmic,shafer2008tutorial,angelopoulos2021gentle}
is a non-parametric calibration method that,
given a value $p$, aims to build a  prediction set with a guarantee that the probability that the correct word is within this set is indeed $p$. Note that this notion of calibration, which is distinct from the way calibration is usually formulated in language modeling settings, \emph{exactly coincides} with the goal of the top-$p$ prediction model. The model-agnostic and distribution-free nature of CP makes it particularly suitable for large neural network models.
We thus applied CP analysis to asses whether the top-$p$ procedure is calibrated and, if needed,  tune it to have the desired probabilistic interpretation. We find that OPT models of different sizes \cite{zhang2022opt} are not calibrated according to the conformal prediction theory, and that calibration shows moderate inverse scaling. Additionally, we show that the degree of calibration varies significantly with the entropy of the model's distribution over the vocabulary. We thus propose a new \textbf{Conformal top-$p$ decoding} algorithm, which ensures that the top-$p$ sampling has a meaningful probabilistic interpretation.

\section{ CP for Language Generation}
In this section, we briefly review the Split Conformal Prediction algorithm \citep{vovk2005algorithmic} and discuss its relevance to language generation models.
Consider a network that classifies an input $x$ into $k$ pre-defined classes. 
The network (softmax layer) output has the mathematical form of a distribution. However,  this does not necessarily mean that it accurately reflects the true class distribution.

Let $(x,y)$ be a test instance and its corresponding class.  
We want to find a small subset of classes (a prediction set) $C(x)\subset\{1,...,k\}$ such that
\begin{equation}
p(y\in C(x)) \ge  1\!-\!\alpha
\end{equation}
where $1\!-\!\alpha\in [0,1]$ is a user-chosen error rate.  
(We use the term $1\!-\!\alpha$ instead of $p$ to comply with CP standard notation).  In words, the probability that the 
set $C(x)$ contains the correct label is at least 
$1\!-\!\alpha$. 
We call this property the marginal coverage since the probability is averaged over all the data points $(x,y)$. 
Denote the prediction set obtained by taking the most probable classes until the total mass just exceeds a value $q$, by $C_{q}(x)$.
Let $\hat{q}\in [0,1]$ be the smallest threshold value that
$ p( y \in C_{\hat{q}}(x))  \ge 1\!-\!\alpha$.
 If $\hat{q} >1\!-\!\alpha$ the model can be viewed as over-confident. If $\hat{q} < 1\!-\!\alpha$ the model can be viewed as under-confident and if $\hat{q}=1\!-\!\alpha$ the model is calibrated in the sense that the probability that the correct label is in the $1\!-\!\alpha$ prediction set is indeed $1\!-\!\alpha$. 

If the model is not calibrated, we can calibrate it using a labeled validation set  
 $(x_1,y_1),...,(x_n,y_n)$. Denote $p_t(i)=p(y_t=i|x_t;\theta)$. Define the \textbf{conformal scores} to be: 
\begin{equation} 
s_t =  \sum_{\{i |  p_t(i) \ge p_t(y_t) \}}  p_t(i) \hspace{0.4cm}
  t=1,...,n
  \label{aps}
  \end{equation}
This CP score is known as the Adaptive Prediction Sets (APS) score, and was first introduced in \citep{romano2020classification}.  
   Note that $ y_t\in C_{s_t}(x_t)$ and $s_t$ is the minimal threshold in which the true class $y_t$ is in a prediction set of $x_t$.
 
We next look for \textbf{a minimal threshold} $\hat{q}$ such that the correct label $y_t$ is included in the prediction set $C_{\hat{q}}(x_t)$ for at least  $(1\!-\!\alpha)n$ points of the validation set. In other words, $\hat{q}$ calibrates  the top-$(1\!-\!\alpha)$ prediction-set on the validation set.
We can easily find $\hat{q}$ by first sorting the $n$ scores $s_1,...,s_n$ 
and then $\hat{q}$ is the $(1\!-\!\alpha)$-quantile of the validation-set scores.
 Once the network is calibrated, if we want to form a prediction set for a new test sample $x$, that contains the true class with probability $(1\!-\!\alpha)$, we use  $C_{\hat{q}}(x)$.
 The CP Calibration procedure for calibrating the  top-$p$ word decoding is summarized in \cref{alg:g_da}.
The conformal prediction theory provides the following guarantee on the threshold $\hat{q}$ \citep{vovk2005algorithmic}.

{\bf Theorem}: Assume  a test point $(x,y)$ and the $n$ validation points are independent and identically distributed (or at least exchangeable).  Let $\hat{q}$ be the $\lceil(n\!+\!1)(1\!-\!\alpha)/n\rceil$-quantile of the  validation set scores.
 Then
 \begin{equation}
1\!-\!\alpha \le p( y\in C_{\hat{q}}(x))  \le 1\!-\!\alpha +\frac{1}{n\!+\!1}.
\label{cptheorem}
\end{equation}
Note that this is a marginal probability over all the test points and is not conditioned on a given input. Exchangeability means that the sequence distribution is not altered by permuting the order of the random variables.

\begin{algorithm}[t]

\begin{algorithmic}
\State {\bf{Input}}: A validation set comprised of next word distributions $p_1,..,p_n$ with the corresponding correct words $y_1,..,y_n$ and a confidence level~$p$.
   \For {$t=1,...,n$} 
   \State $s_t =  \sum_{\{i |  p_t(i) \ge p_t(y_t) \}}  p_t(i) $
   \EndFor
   \State Define $\hat{q}$ to be the $\lceil(n\!+\!1)p/n\rceil$-quantile of $  
     \{s_1,...,s_n\}$.
   \State {\bf{Output:}} Use top-$\hat{q}$ decoding to guarantee that the probability that the correct word is in the top-$\hat{q}$ prediction set is at least $p$.  
   \caption{ CP Calibration of the Top-$p$ decoding}\label{alg:g_da}
   \end{algorithmic}
   
\end{algorithm}

In this study, we aim to apply the conformal prediction framework to language generation models  to analyze the prediction sets used for sampling the next word. 
The joint distribution of words in a text is neither IID nor exchangeable, since the words are correlated and the order of the words in a sentence is significant.
A recent study \citep{oliveira2022split}  showed that applying the usual CP algorithm to a stationary $\beta$-mixing process (rather than an exchangeable one) results in a guaranteed  coverage level of $1\!-\!\alpha\!-\!\eta$, where $\eta$ depends on the mixing properties of the process and is theoretically hard to know, or bound.  
Roughly speaking, $\beta$-mixing processes are stochastic processes in which far-away points are approximately independent in a quantifiable manner. In all the examples they checked, the authors assessed that the additional penalty incurred by using CP with stationary $\beta$-mixing processes was virtually insignificant.
\citet{manning1999foundations} argue that even though not quite correct, natural language can be modeled as stationary, ergodic
processes. \citet{khandelwal2018sharp} 
 showed that the  LSTM language
model’s memory is empirically bounded at roughly 200 words and thus the model can be viewed as an aperiodic recurrent (and therefore $\beta$-mixing)  Markov chain.
It is reasonable to assume that human language and transformer-based language models can also be modeled  as $\beta$-mixing processes.
 Hence, applying  CP to language generation models yields meaningful results (at least qualitatively).

\section{Experiments}
\label{sec:experiments}





In this section, we apply the conformal prediction calibration method to analyze the calibration status of the top-$p$ nucleus sampling.

\begin{figure}[t]
\centering
\includegraphics[width=\linewidth]{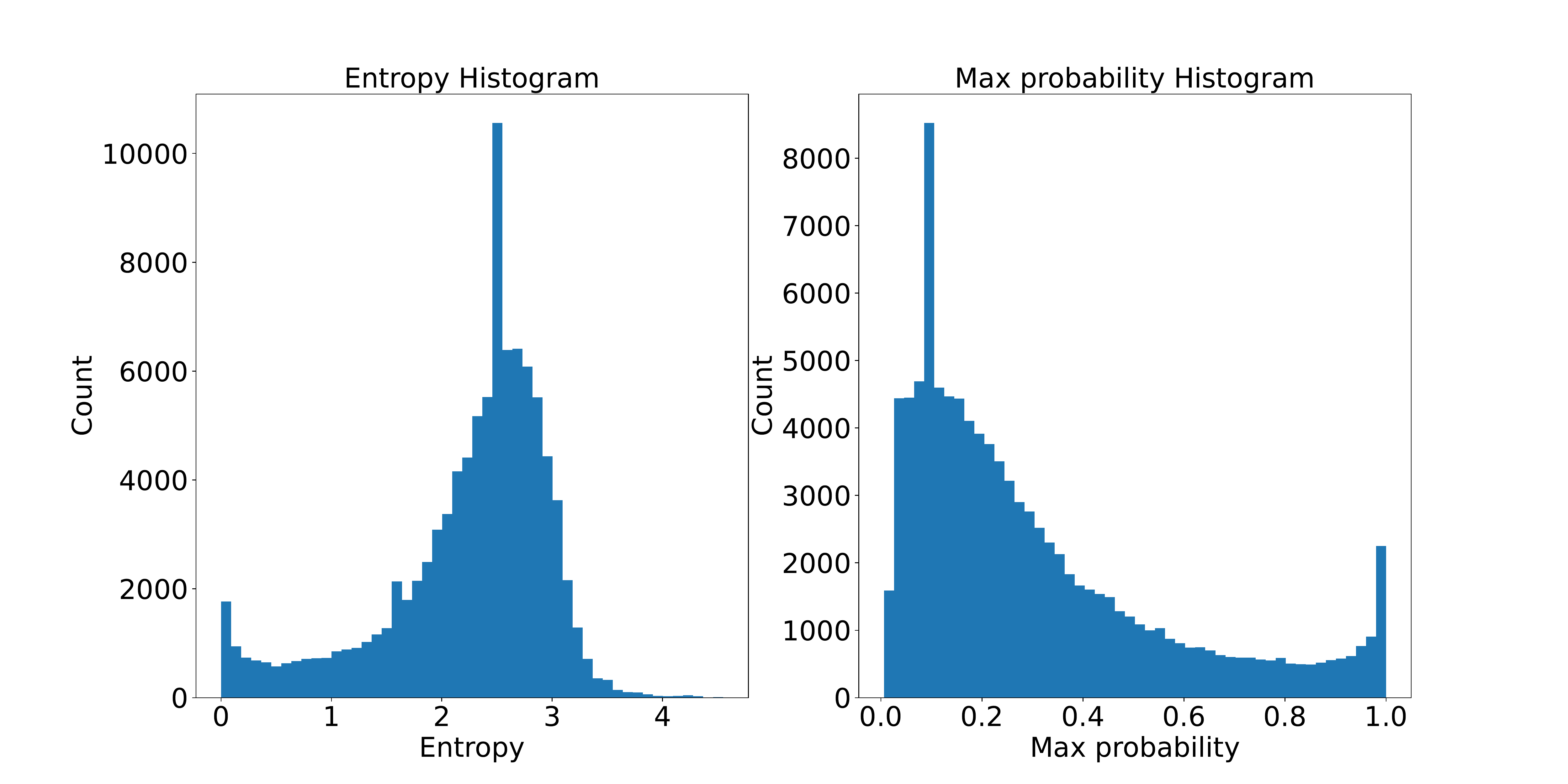}
\caption{Histograms of entropy of the output probability distribution for the OPT350M model.}
\label{fig:entropy-hist-opt-350}
\end{figure}

\begin{figure}[h]
\centering
\includegraphics[width=\linewidth]{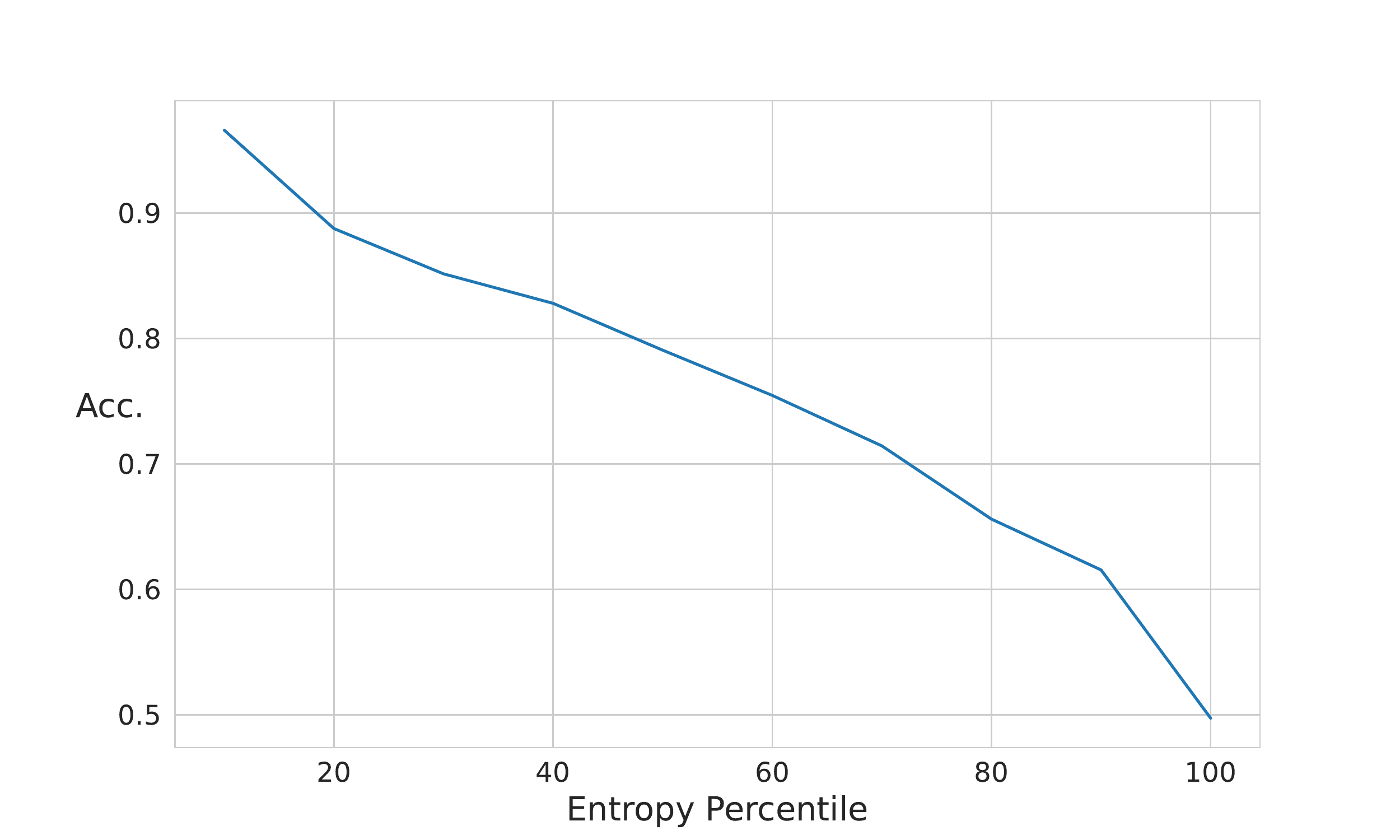}
\caption{Effective accuracy when using nucleus sampling with $p=0.9$, for different entropy percentiles, for the OPT350M model. 
}
\label{fig:acc-vs-p-per-entropy-opt-350}
\end{figure}

\begin{figure}[h]
\centering
\includegraphics[width= 6.8cm]{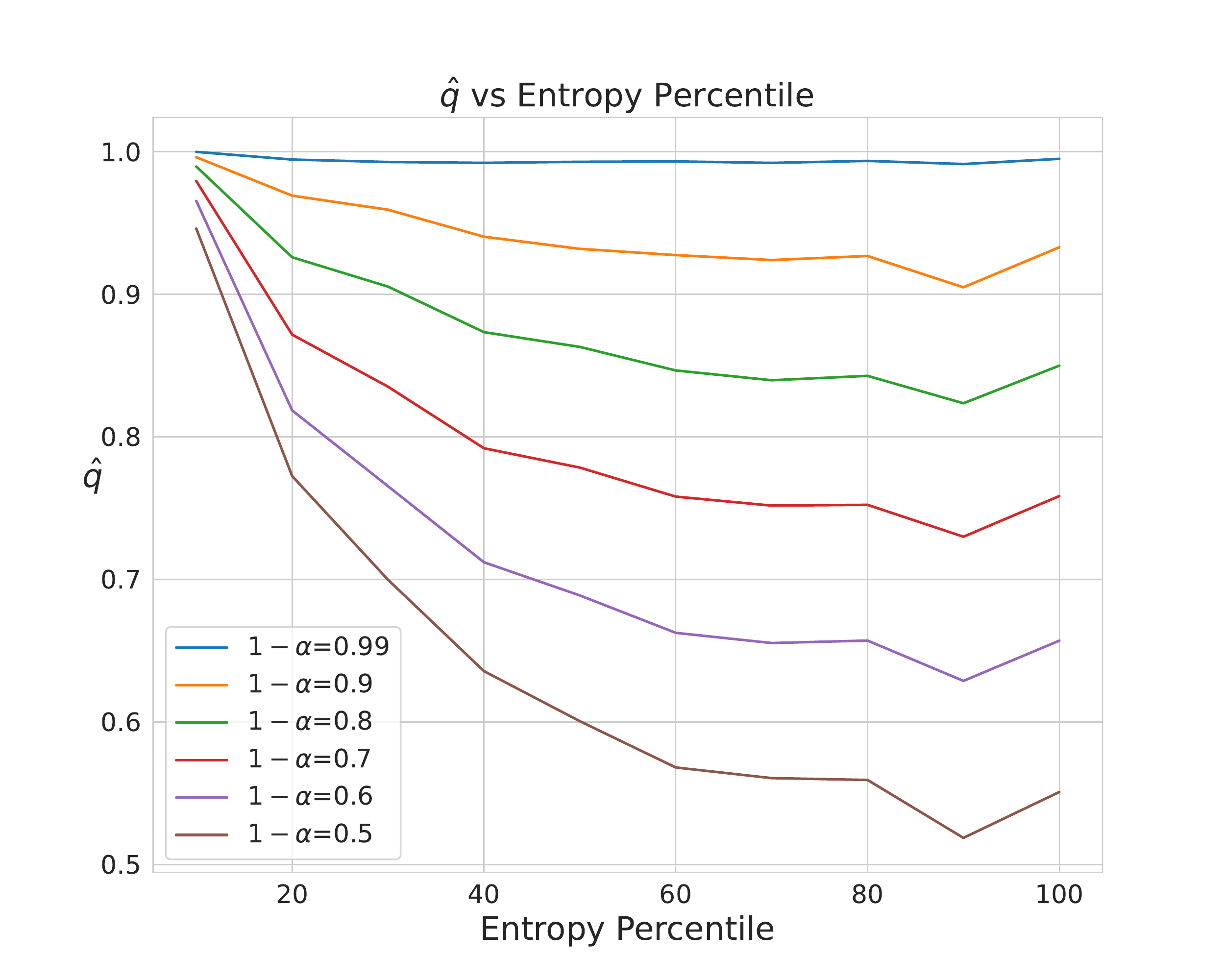}
\caption{$\hat{q}$ threshold scores when calibration is performed over the examples belonging to each entropy percentile separately.}
\label{fig:q-vs-alpha-per-entropy-opt350}
\end{figure}

{\bf{Setup.}}
We experimented with variants---from 125M parameters up to 30B parameters---of OPT \cite{zhang2022opt}, a left-to-right language model. We ran the models on 10,000 English Wikipedia sentences\footnote{\href{https://huggingface.co/datasets/wikipedia}{https://huggingface.co/datasets/wikipedia}}, and collected the distribution of the vocabulary over each token in each sentence, resulting in a total of 245,923 distributions. 
The distribution of the entropy values, as well as the maximum probability, was far from being uniform (\cref{fig:entropy-hist-opt-350}). We sorted all the instances by entropy, and calibrated the examples belonging to each equally-sized percentile independently (from 0-10\% to 90-100\%). The patterns are highly similar across models. We report results on the 350M parameters model unless specified otherwise. We use Nvidia 2080TI GPUs. 
 
\paragraph{Dependency of the confidence on the entropy.}  First, we evaluated the confidence scores of a standard nucleus sampling scheme. We chose $p=0.9$ (a commonly used value) and recorded the effective confidence, i.e., the proportion of cases where the correct word was indeed in the top-$p$ prediction set. 
\cref{fig:acc-vs-p-per-entropy-opt-350} shows the effective confidence for the predictions belonging to different percentiles of entropy. The results  indicated that setting $p=0.9$ did not translate to a prediction set that contained the correct token in 90\% of the cases, motivating our calibrated decoding. In \cref{fig:q-vs-alpha-per-entropy-opt350}, we show the per-entropy CP calibration results, for 10 entropy bins corresponding to percentiles. While the model was always overconfident, the level of overconfidence decreases with the entropy percentile. In other words,  when the model is apparently the most certain---as reflected in low entropy values---it is most overconfident. Note that in the case of low entropy the single highest probability can be more than 0.9. Hence, there is no way to calibrate the prediction set by changing its size. 
In particular, we found that the model is overconfident when the gold token is a function word: it tends to allocate high probability to a small set of function words, while the true distribution is more varied.

\begin{figure}[t]
\centering
\includegraphics[width=\linewidth]{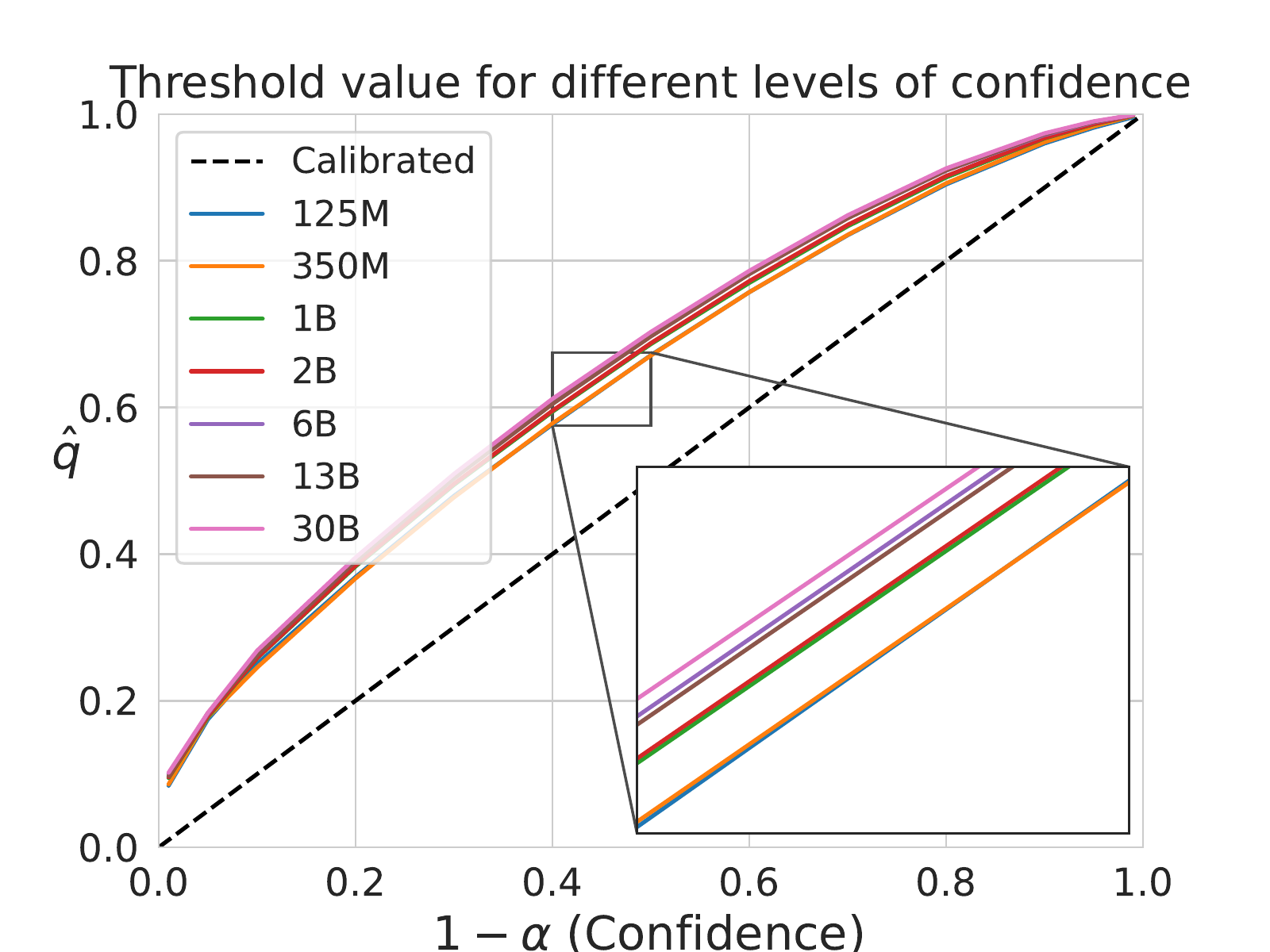}
\caption{$\hat{q}$ threshold values needed to ensure a confidence of 1-$\alpha$. The OPT models show slight inverse scaling with respect to calibration.}
\label{fig:scaling}
\end{figure}

\paragraph{Calibration and scale.}  \cref{fig:scaling} presents the conformal threshold values $\hat{q}$ versus desired confidence ($1\!-\!\alpha$), when calibration is performed over the entire validation set (without partition to entropy bins). As shown, for all confidence levels, the threshold $\hat{q}$ needed to ensure that the correct word is included within the prediction set is larger than the confidence level itself (the $y=x$ dashed line). This indicates that the model is \emph{overconfident}. 
\cref{fig:scaling} also shows the dependency of calibration on the scale. Scaling language models has been shown to induce the emergence of new abilities, such as in-context learning \cite{brown2020language}. Empirical power laws were shown to predict performance in a different task as a function of scale \cite{kaplan2020scaling, wei2022emergent}, where models usually show improved performance with scale. Here, we find \emph{inverse scaling} \cite{wei2022inverse}, where calibration moderately deteriorates with model scale.

\paragraph{Generation.}
How does conformal $p$ sampling affect generation? we use the 350M model to compare the quality of generation of conformal $p$ sampling with the natural baseline of $p$ sampling. 
We generate continuations to 1,000 prompts of size 35 words from the OpenWebText dataset \footnote{\href{https://github.com/jcpeterson/openwebtext}{https://github.com/jcpeterson/openwebtext}}. We generate up to length 200 tokens, and compare conformal $p=0.9$ prediction (setting $1-\alpha=0.9$) with conventional $p=0.9$ sampling.\footnote{We make the generations available at \href{https://anonymous.4open.science/r/CP-40CD/gens_ours.pickle}{this} link.} Following \cref{fig:q-vs-alpha-per-entropy-opt350}, when applying our method, we calculate the entropy of the output distribution over each token, and dynamically set the threshold $p$ for each token prediction, according to the threshold value $\hat{q}$ that fits this entropy percentile. This ensures that the \emph{true probability} of the token to be included within the prediction set (according to the training set used for calibration) is $0.9$. 

We evaluate the quality of the generation using MAUVE \cite{pillutla2021mauve} and BERTScore \cite{zhang2019bertscore}.\footnote{Default HuggingFace v4.22.0 Parameters were used.} MAUVE score is $0.933$ for conformal-$p$ sampling, and $0.0.920$ for conventional $p$ sampling. As for BERTScore, the $F1$ score is $0.840$ for conformal-$p$ sampling, and $0.843$ for conventional $p$ sampling. These results indicate that conformal-$p$ sampling is performing similarly to conventional $p$ sampling. 

\paragraph{Applicability of CP to non IID data}  Conformal prediction theory assumes IID, while we build on the model outputs distributions over consecutive tokens in the same sentence, which are of course highly dependent. We repeated the per-entropy-bin calibration process when uniformly sampling a \emph{single} token per sentence, thus (almost) satisfying the independence assumption. The results were similar to \cref{fig:q-vs-alpha-per-entropy-opt350} and in that case, \cref{cptheorem}) is applicable.

\section{Conclusions}
To conclude, in this study we apply the notion of calibration by conformal prediction to calibrate the top-$p$ nucleus sampling as a function of the next word distribution entropy and thus made the top-$p$ decoding policy consistent.
The same analysis and calibration can also be applied to other commonly used decoding methods, such as variants of top-$p$ \citep{Meister2022} and truncation sampling \citep{Hewitt2022}.

\section*{Limitations}
We calibrated OPT models based on Wikipedia data. Future work should apply calibration procedure to a wider range of datasets, to check whether our results generalize to different domains. Additionally, we limited our evaluation to entropy as a measure of uncertainty and did not explore other measures. Finally, we aimed at validating the calibration status of commonly used LMs. Future work should thoroughly evaluate the impact of the calibration status on different facets of generation quality, as text generation is one of the main use-cases of large LMs.

\section*{Ethics Statement}
We do not foresee ethical issues with this work.

\section*{Acknowledgements}
This project received funding from the Europoean Research Council (ERC) under the Europoean Union's Horizon 2020 research and innovation programme, grant agreement No. 802774 (iEXTRACT). Shauli Ravfogel is grateful to be supported by the Bloomberg Data Science Ph.D. Fellowship.

\bibliography{anthology,custom}
\bibliographystyle{acl_natbib}

\end{document}